\documentclass{article}
\usepackage{ijcai22}

\usepackage{times}

\usepackage{soul}
\usepackage{url}
\usepackage[hidelinks]{hyperref}
\usepackage[utf8]{inputenc}
\usepackage[small]{caption}
\usepackage{graphicx}
\usepackage{amsmath, bm}
\usepackage{booktabs}
\usepackage{color}
\usepackage{xcolor}
\hypersetup{
    colorlinks,
    linkcolor={blue!50!black},
    citecolor={blue!50!black},
    urlcolor={blue!80!black}
}
\usepackage{natbib}
\hypersetup{
    colorlinks,
    linkcolor={blue!50!black},
    citecolor={blue!50!black},
    urlcolor={blue!80!black}
}
\usepackage[linesnumbered, ruled]{algorithm2e}
\usepackage{relsize}

\usepackage{multirow}

\title{An end-to-end predict-then-optimize clustering method for intelligent assignment problems in express systems}

\author{
Jinlei Zhang$^1$\and
Ergang Shan$^2$\and
Lixia Wu$^{2}$\and
Lixing Yang$^1$ \footnote{Corresponding Author} \and
Ziyou Gao$^3$\And
Haoyuan Hu$^2$\\
\affiliations
$^1$State Key Laboratory of Rail Traffic Control and Safety, Beijing Jiaotong University\\
$^2$Zhejiang Cainiao Supply Chain Management Co. Ltd\\
$^3$School of Traffic and Transportation, Beijing Jiaotong University\\
\emails
zhangjinlei@bjtu.edu.cn,
\{ergang.se, wallace.wulx\}@cainiao.com,
\{lxyang, zygao\}@bjtu.edu.cn,
\{haoyuan.huhy\}@cainiao.com
}

\begin{document}
\pagestyle{plain}
\pagenumbering{arabic}

\maketitle

\begin{abstract}
Express systems play important roles in modern major cities. Couriers serving for the express system pick up packages in certain areas of interest (AOI) during a specific time. However, future pick-up requests vary significantly with time. While the assignment results are generally static without changing with time. Using the historical pick-up request number to conduct AOI assignment (or pick-up request assignment) for couriers is thus unreasonable. Moreover, even we can first predict future pick-up requests and then use the prediction results to conduct the assignments, this kind of two-stage method is also impractical and trivial, and exists some drawbacks, such as the best prediction results might not ensure the best clustering results. To solve these problems, we put forward an intelligent end-to-end predict-then-optimize clustering method to simultaneously predict the future pick-up requests of AOIs and assign AOIs to couriers by clustering. At first, we propose a deep learning-based prediction model to predict order numbers on AOIs. Then a differential constrained \textit{K}-means clustering method is introduced to cluster AOIs based on the prediction results. We finally propose a one-stage end-to-end predict-then-optimize clustering method to assign AOIs to couriers reasonably, dynamically, and intelligently. Results show that this kind of one-stage predict-then-optimize method is beneficial to improve the performance of optimization results, namely the clustering results. This study can provide critical experiences for predict-and-optimize related tasks and intelligent assignment problems in express systems.
\end{abstract}


\section{Introduction}
Express systems play an important role in the daily life of urban citizens. Large amounts of packages remain to be delivered every day. Customers generally place a pick-up order with their private information, such as the name, phone number, location, etc, in the Express APP \citep{li2020cooperative}. Couriers can receive the pick-up request in real-time and then take action to pick the package up to help customers send the package out. During this process, the current operation mode for the express company is static and fixed. Several critical issues are to be noted as follows. 
\begin{itemize}
    \item The administrator divides urban areas into several irregular regions according to the road network, which is called the area of interest (AOI).
    \item All pick-up requests from one AOI will be assigned to only one courier. That means the courier will be responsible for all of the pick-up requests from this AOI.
    \item One single AOI will be assigned to at most one courier.
    \item One courier can serve for several AOIs.
    \item The assignment results are generally fixed without changing with time.
\end{itemize}

\begin{figure}
    \centering
    \includegraphics[width=0.4\textwidth]{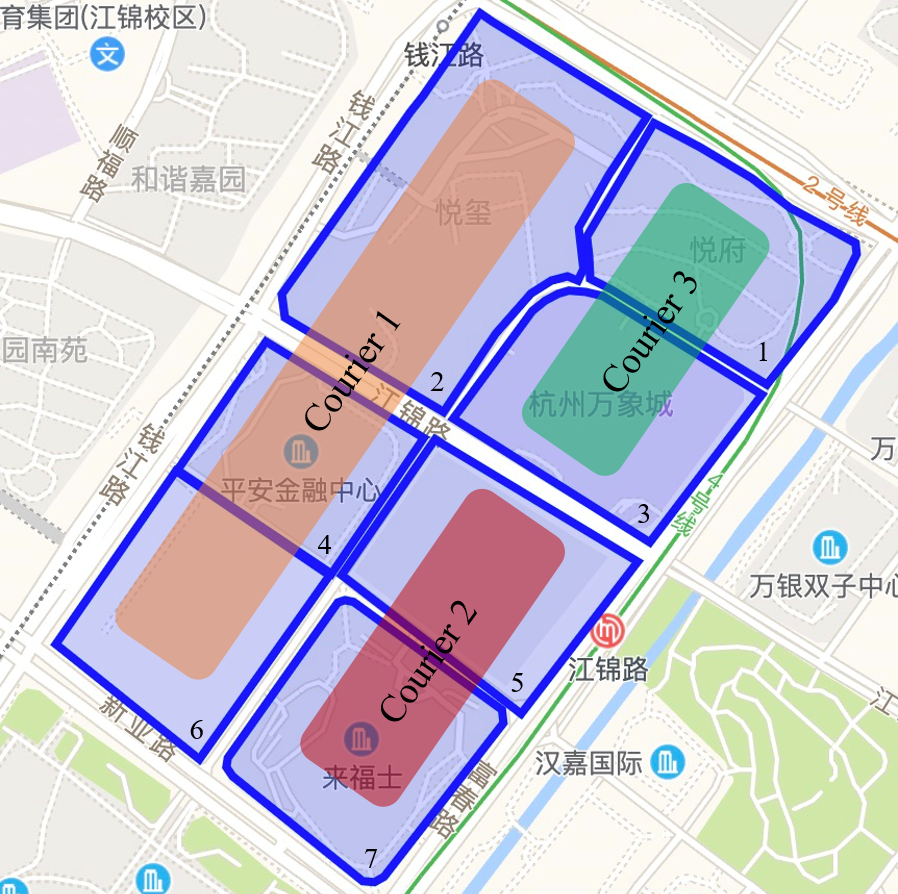}
    \caption{The assignment diagram of AOIs and couriers}
    \label{fig:AOI}
\end{figure}

The assignment diagram of AOIs and couriers are shown in Figure~\ref{fig:AOI}. As is shown, There are totally seven AOIs in the diagram. They are assigned to three couriers according to the administrator's experiences considering the location of the AOI and the order number of pick-up requests from the AOI. Courier 1 is responsible for 2, 4, 6. Courier 2 is responsible for 5, 7. Courier 3 is responsible for 1, 3. However, there are several issues and challenges to be considered during the AOI assignment process.

\textbf{The AOI partition results are static and fixed.} In the express system, the administrator generally partitions urban areas into AOIs according to the road network. Once the partition is given, they generally remain unchanged because the manual partition process consumes substantial cost and time.

\textbf{The AOI assignment results are static and fixed.} In the express system, the administrator generally assigns AOIs to couriers according to the location of the AOI and the historical order number of the AOI to ensure that the total order number that a courier is responsible for is approximately equal. So that all couriers can get approximately equal income.

\begin{figure}
    \centering
    \includegraphics[width=0.45\textwidth]{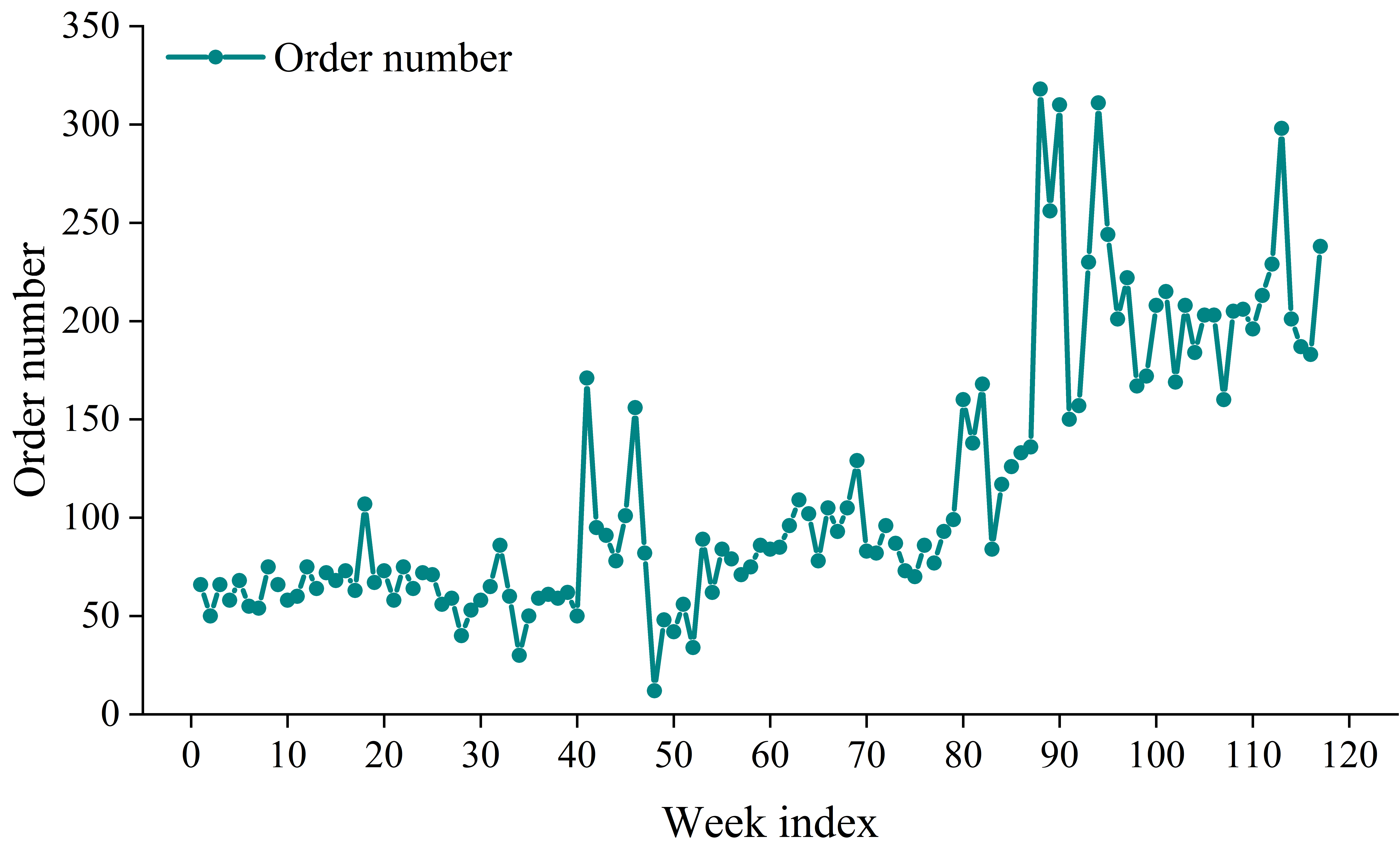}
    \caption{The variation of order number with week}
    \label{fig:order}
\end{figure}

\textbf{The order number of the AOI changes largely with time.} Affected by tremendous factors, such as festivals, sales promotion, etc, the order number of each AOI often changes largely with time as shown in Figure~\ref{fig:order} \citep{li2019efficient}. However, the AOIs that each courier is responsible for are fixed, which causes the total orders each courier is responsible for also change largely. Moreover, the future order number of each AOI is significantly dynamic. Experience-oriented AOI assignment exhibits great irrationality.

In terms of these issues, one of the feasible solutions is to dynamically cluster AOIs (cluster results can be treated as assignment results for couriers) according to the future order numbers of each AOI. Therefore, we can first predict the order number of each AOI, and then conduct the clustering according to the prediction results. During the clustering process, constraints of the total order number of each cluster must be guaranteed to ensure that all couriers are responsible for approximately equal orders. This is a typical "predict-then-optimize" problem \citep{elmachtoub2021smart}. However, this kind of two-stage solution presents several limitations. \textit{\textbf{First}}, the criteria by which we train the prediction model differ from the ultimate criteria on which we evaluate the clustering model \citep{donti2017taskbased}. For example, the loss function of the prediction model is generally mean square error (MSE), while the criteria for evaluating the clustering results is generally the modularity. Our ultimate goal is to obtain the best clustering result (the highest modularity) under some constraints rather than the best prediction results (the smallest MSE). \textit{\textbf{Second}}, the optimal prediction results might not guarantee the optimal clustering results. That is, the clustering results might be sub-optimal provided that the prediction model is optimal \citep{wilder2019end}. In other words, the smallest MSE cannot guarantee the highest modularity. It seems somewhat counterintuitive because someone regards that the best prediction results would of course guarantee optimal optimization results. However, it should be noted that there also exist errors even with the best prediction model. Therefore, it is necessary to trade off between the objective of the optimal clustering results and the errors of the  prediction results \citep{wilder2019end}. \textit{\textbf{Third}}, from the perspective of real-world application, it is impractical and trivial  to implement such a two-stage solution.

In this study, we propose an intelligent AOI assignment method in express systems, namely an end-to-end predict-then-optimize (\textbf{PTO}) clustering method (called \textbf{PTOCluster}). In this framework, we first propose a graph convolutional neural network (GCN)-based \citep{kipf2016semi} prediction model to predict the total order number of each AOI. Then, an optimization model subject to a specific objective and constraints with the predicted order number as parameters is put forward to obtain the clustering results. We propose a differential \textit{K}-means with constraints as the optimization model. So that it can be treated as a learnable optimization layer following the prediction layer, thus comprising the PTOCluster framework. In the PTOCluster framework, our ultimate goal is to obtain the optimal clustering result in the context of the optimization model and prediction results. Therefore, the loss function of the PTOCluster is the modularity that evaluates the clustering results rather than the MSE that evaluate the prediction results. The contributions can be summarized as follows.
\begin{itemize}
    \item We propose a smart PTOCluster framework for the intelligent and dynamic AOI assignment in express systems. This framework can automatically trade off between the objective of the optimal clustering results and the errors of the  prediction results.
    \item A GCN-based deep learning prediction model used to predict the order number of each AOI is proposed.
    \item In the context of dynamic prediction results, a differential \textit{K}-means with constraints is proposed as an optimization model and a learnable optimization layer for the first time.
    \item The PTOCluster framework is tested on two real-world datasets from the express system and results show its superiority over the two-stage model.
\end{itemize}

The remainder is organized as follows. Related works are presented in Section~\ref{Relatedworks}. We define the problem in Section~\ref{Problem}. The methodology is described in Section~\ref{Methodology}. The Experimental results are discussed in Section~\ref{Experimental}. We conclude this study in Section~\ref{Conclusion}.

\section{Related works} \label{Relatedworks}
In fact, there are few studies exploring the dynamic AOI assignment problem. Therefore, we only review the literature relevant to order prediction, \textit{K}-means clustering, and end-to-end PTO problem.

\textbf{Order prediction in express systems:} Order prediction is a kind of time series prediction task. In terms of time series prediction, relevant prediction models experience a long history \citep{Zhang2021Deep}. Early models are statistic-based, such as the historical average method, least square method, and autoregressive integrated moving average method. To improve the prediction accuracy, machine learning-based models are proposed, such as support vector machine, random forest regression, and \textit{K}-nearest neighbor. With the development of deep learning, tremendous deep learning-based models have been applied to time series prediction tasks because of their powerful spatiotemporal learning ability, such as recurrent neural network (RNN), convolutional neural network (CNN), and graph convolutional neural network (GCN). With respect to the order prediction in express systems, because there are many factors affecting the order number, the variation of order number often fluctuates significantly as shown in Figure~\ref{fig:order}. Therefore, it is a critically hard problem to conduct the order prediction. Only a few studies explored similar problems. The closest of this study is \cite{ren2021deepexpress}. They proposed a deep learning architecture based on the long short-term memory to predict the number of next daily parcels in some regions. Our study differs from theirs in that we use GCN to capture the topological information between regions \citep{kipf2016semi}. Moreover, the GCN is faster and more efficient than RNN-related models during the training process. This characteristic enables the GCN right suitable for the PTOCluster problem. This is  because the prediction model and optimization model will be implemented sequentially and iteratively when training the PTOCluster, which requires both of the two models to be efficient enough.

\textbf{\textit{K}-means clustering method:} \textit{K}-means was initially proposed as a conventional, efficient, and unsupervised cluster method \citep{hartigan1979algorithm}. In recent years, \cite{wilder2019end} approximated the \textit{K}-means as a differential version so that is could be embedded into the deep learning framework as a neural network layer. However, there is no constraint in their differential \textit{K}-means. \cite{wagstaff2001constrained,bradley2000constrained} proposed a constrained version of \textit{K}-means. \cite{le2018binary, agoston2021mixed, moreno2020hybrid} treated \textit{K}-means as an integer programming problem to solve. However, they cannot be differentiated because the objective and constraints are not continuous. In this study, we propose a differential \textit{K}-means especially with constraints. This version of \textit{K}-means is expressed as a kind of stochastic programming problem with the prediction results as parameters. The clustering results are soft-label solutions. A hard maximum to the soft-label solutions will be applied to obtain the final clustering results \citep{wilder2019end}.

\textbf{End-to-end predict-then-optimize problem:} \textbf{In terms of the PTO problem,} there are two kinds of solutions, one is two-stage solutions and the other is one-stage, or end-to-end, solutions. For two-stage solutions as mentioned in the introduction section, an optimal prediction result might not ensure an optimal optimization result and the two-stage solution is impractical and trivial for real-world applications. For one-stage solutions, recent years have seen a dramatic increase in the number of end-to-end PTO researches. Most of them seek to build an end-to-end framework to solve PTO problems so as to obtain optimal optimization results \citep{puppels2020investigation, yan2021surrogate}. Generally speaking, the prediction part is a machine learning- or deep learning-based model, such as the fully connected layer, the GCN layer, etc \citep{donti2017taskbased, wilder2019end}. The optimization part is an optimization model treated as a differential optimization layer, including linear programming \citep{wilder2019melding}, quadratic programming or more general convex optimization \citep{amos2017optnet, diamond2016cvxpy, agrawal2019differentiable}, mixed integer programming \citep{paulus2020fit, elmachtoub2020decision, elmachtoub2021smart}, combinatorial optimization \citep{mandi2020smart, wilder2019melding, wilder2019end}, stochastic optimization \citep{wilder2019end}, etc. Among these problems, some are with constraints while some are without constraints, some are with discrete decision variables while some are with continuous decision variables.

\section{Problem definition} \label{Problem}
The goal of this study is to dynamically assign AOIs to couriers by considering the future pick-up order numbers and geographic locations of AOIs. Therefore, we will first predict the future pick-up order number of AOIs, and then cluster the AOIs into several categories under some constraints using the predicted order number and geographic locations of AOIs. 

As for the prediction process, we define the area with many AOIs as a graph $G=(V, E, A)$, where $V=(V_1, V_2, V_3, ..., V_n)$ are the centers of AOIs, $E$ are the edges between the centers of adjacent AOIs, and $\bm{A} \in \bm{R}^{n \times n}$ is the adjacent matrix of all AOIs with all elements being one or zero. For example, the adjacent matrix of Figure~\ref{fig:AOI} ($\bm{A} \in \bm{R}^{7 \times 7}$) is Equation~\eqref{eq1}. The feature matrix of AOIs is $\bm{X} \in \bm{R}^{n \times m}=(\bm{X}_t,\bm{X}_{t-1},\bm{X}_{t-2},...,\bm{X}_{t-m+1})$ where $n$ is the number of AOIs, $m$ are the historical time steps used to predict the order series in the time interval $t+1$, ($\bm{X}_t \in \bm{R}^{n \times 1}$) is a column vector denoting the order series in a specific time interval. Therefore, the prediction process can be defined as Equation~\eqref{eq2}, in which the $\bm{Y}$ is the predicted future order numbers of AOIs. When only conducting the prediction, the loss function is generally the Mean Squared Error (MSE).

\begin{equation}       
\bm{A}= \left(                 
  \begin{array}{ccccccc}   
    0 & 1 & 1 & 0 & 0 & 0 & 0 \\  
    1 & 0 & 1 & 1 & 0 & 0 & 0 \\  
    1 & 1 & 0 & 0 & 1 & 0 & 0 \\  
    0 & 1 & 0 & 0 & 1 & 1 & 0 \\  
    0 & 0 & 1 & 1 & 0 & 0 & 1 \\  
    0 & 0 & 0 & 1 & 0 & 0 & 1 \\  
    0 & 0 & 0 & 0 & 1 & 1 & 0 \\  
  \end{array}
\right)                 
\label{eq1}
\end{equation}

\begin{equation}      
\bm{Y}=f_1(\bm{X}_t,\bm{X}_{t-1},\bm{X}_{t-2},...,\bm{X}_{t-m+1}; \bm{A})
\label{eq2}
\end{equation} 

As for the clustering process, we propose a differential constrained \textit{K}-means as an optimization layer. This layer uses the predicted $\bm{Y}$ together with the geographic location information of AOI centers as inputs, and the clustering results as outputs, as shown in Equation~\eqref{eq2}. We will first obtain a soft solution and then round it to a discrete solution. The criteria used to evaluate the clustering result is modularity \citep{newman2006modularity, wilder2019end}. The larger the indicator, the better the clustering result. Therefore, we apply the opposite number of the modularity as the loss function of this optimization layer.

\begin{equation} 
\bm{Z}=f_2(\bm{Y},\bm{\alpha}_i)
\label{eq3}
\end{equation} 
where $\bm{Y}$ is the final clustering result, $\bm{\alpha_i}=(\bm{\lambda}_i, \bm{\varphi}_i)$ are the longitude and latitude of AOI centers.

As for the PTO framework ($ \bm{X \to Y \to Z} $), the input is historical order numbers of AOIs and the output is future clustering results, namely the assignment results, as shown in Equation~\eqref{eq4}. The loss function for this end-to-end framework is modularity rather than the MSE. That means we use the optimal solution of the optimization problem to guide the training process of prediction layers so that we can obtain the best clustering result rather than the best prediction results. 
\begin{equation} 
    \begin{split}
        \bm{Z}&=F(f_1,f_2)\\
        &=F(\bm{X}_t,\bm{X}_{t-1},\bm{X}_{t-2},...,\bm{X}_{t-m+1};\bm{A}; \bm{\theta})\\
        &=F(\bm{X};\bm{A}; \bm{\theta})
    \end{split}
\label{eq4}
\end{equation} 
where $\theta$ denotes all of the parameters in the prediction layer.

\section{Methodology} \label{Methodology}
In this section, we will present the methodology, including the PTOCluster framework, the prediction model, and the optimization or clustering model.
\subsection{PTOCluster framework}
\begin{figure*}[t]
    \centering
    \includegraphics[width=0.7\textwidth]{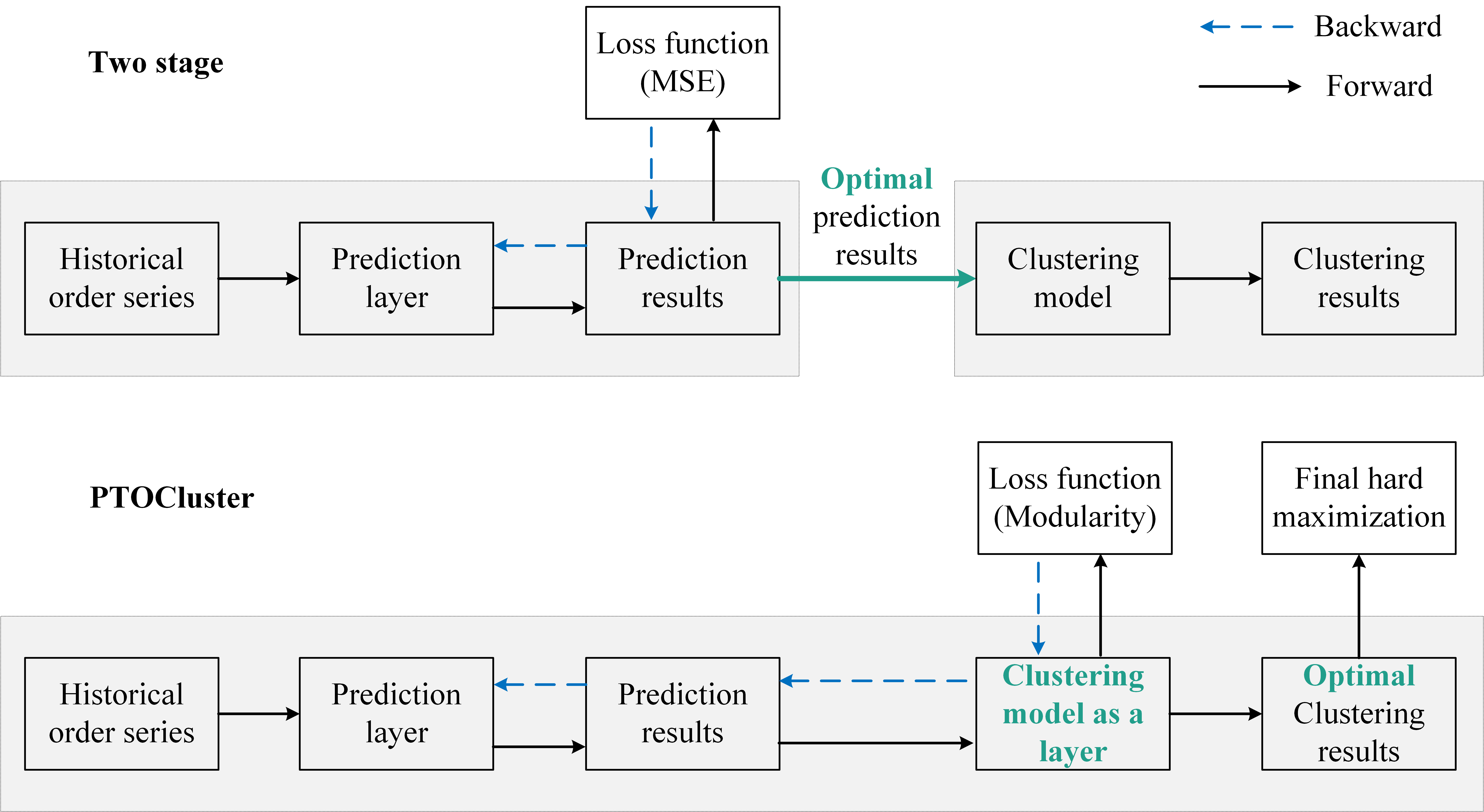}
    \caption{The two-stage approach and the proposed PTOCluster framework}
    \label{fig:framework}
\end{figure*}

Figure~\ref{fig:framework} shows the two-stage approach and the proposed PTOCluster framework. We also highlight the differences between the two approaches. For the two-stage approach, the MSE is generally used to train the prediction model. When the \textit{\textbf{optimal prediction results}} are obtained, they are input into the optimization model, namely the clustering model, to obtain the clustering results leveraging some solvers. For the proposed PTOCluster framework, rather than use the intermediary MSE loss function, we apply the modularity, which can evaluate the clustering results, as the loss function to train the prediction model as well as simultaneously obtain the \textit{\textbf{optimal clustering results}}.

\subsection{Forward pass}
During the forward pass, data passes through the prediction layer and the clustering optimization layer. When the training process ends, the optimal clustering results, namely the future AOI assignment results, can be obtained. Therefore, we will present the details of the two models respectively. 

\subsubsection{\textbf{(1) GCN-based prediction model}}

\begin{figure*}[t]
    \centering
    \includegraphics[width=1\textwidth]{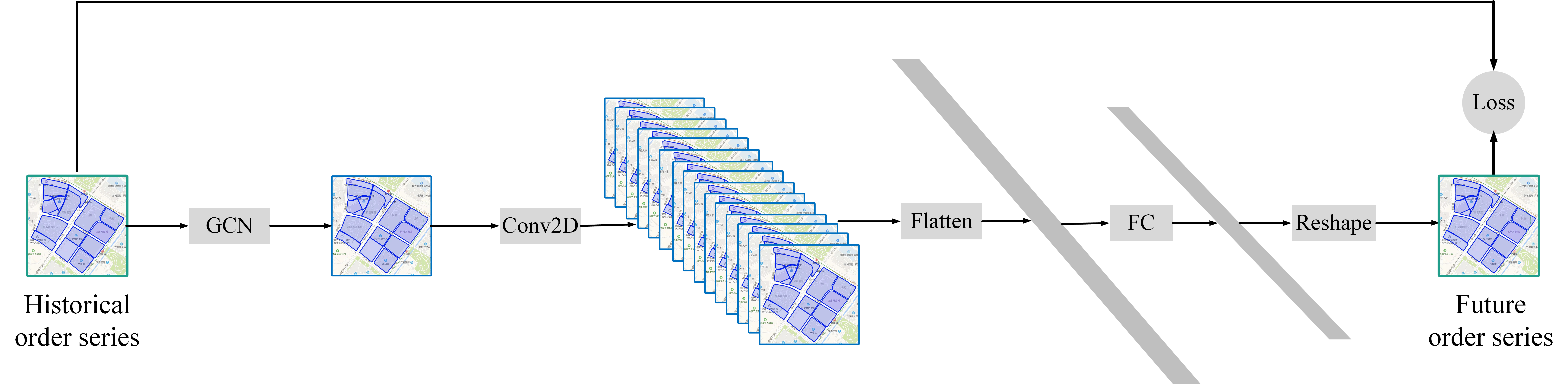}
    \caption{GCN-based prediction framework}
    \label{fig:predictionmodel}
\end{figure*}

In terms of the prediction layer, we propose a GCN-based deep learning architecture for the order prediction as shown in Figure~\ref{fig:predictionmodel}. The GCN is followed by a 2D CNN layer and three fully connected (FC) layers. As for the GCN, we utilize the first-order approximation version proposed by \cite{kipf2016semi}, as shown in Equation~\eqref{eq5}.
\begin{equation} 
    \bm{Y}=\sigma(\tilde{\bm{D}}^{-\frac{1}{2}} \tilde{\bm{A}} \tilde{\bm{D}}^{-\frac{1}{2}}\bm{XW} + \bm{b})
    \label{eq5}
\end{equation} 
where $\bm{A}$ is the adjacent matrix, $\tilde{\bm{A}}=\bm{A}+\bm{I}$, $\bm{I}$ is the identity matrix, $\tilde{\bm{D}}$ is the diagonal node degree matrix of $\tilde{\bm{A}}$, $\bm{X}$is the feature matrix, $\bm{W}$is the weight matrix, $\bm{b}$ is the bias.

\subsubsection{\textbf{(2) Clustering optimization model}}

When dynamically assigning AOIs to couriers, several issues should be considered. First, the iteratively predicted order number can be treated as weighted parameters of AOIs, thus ensuring  the total order number that a courier is responsible for is approximately equal so that all couriers can get approximately equal income. Second, the geographic location information of AOI centers should be included in the model so that couriers are responsible for adjacent AOIs.

To achieve these goals, we propose a differential constrained \textit{K}-means as an optimization layer, as shown in Equation~\eqref{eq6}.

\begin{align}
	\min_{z_{i,p}} &\quad \sum_{i=1}^{n} \sum_{p=1}^{m} y_i \cdot z_{i,p} \cdot ||\alpha_i - C_p|| \label{eq6}\\
    s.t. &\quad \sum_{i=1}^{n}y_i \cdot z_{i,p} \leq a_{u} && p=1, 2,\cdots,m  \label{eq7}\\
         &\quad \sum_{i=1}^{n}y_i \cdot z_{i,p} \geq b_{l} && i=1, 2,\cdots,n  \label{eq8}\\
         &\quad \sum_{p=1}^{m}z_{i,p} =1 && i=1, 2,\cdots,n  \label{eq9}\\
         &\quad z_{i,p} \in \{0,1\} &&i=1, 2,\cdots,n     \label{eq10}\\
         &\quad &&p=1, 2,\cdots,m \nonumber 
\end{align}
where 
\begin{itemize}
    \item $i$ is the index of AOI and there are totally $n$ AOIs.
    \item $y_i$ is the predicted order number of AOI $i$.
    \item $p$ is the index of the clustering number and there are totally $m$ clusters.
    \item $z_{i, p}$ is the decision variable indicating whether the AOI $i$ belongs to cluster $p$.
    \item $\alpha_i=(\lambda_i, \varphi_i)$ are the longitude and latitude of centers of AOI $i$.
    \item $C_p = \frac{\sum_i y_i z_{i,p}\alpha_i}{\sum_i y_i z_{i,p}}$ is the clustering center of cluster $p$.
    \item $a_l$ and $b_u$ are the lower bound and upper bound of the order number that a courier is responsible for, respectively.
\end{itemize}

The objective is to minimize the order number-weighted distance between the AOI centers and the clustering centers. The first and second constraints are to ensure the total order number that a courier is responsible for is approximately equal, namely within a reasonable interval $[-b_{l}, a_{u}]$. The third constraint ensures that an AOI can only be assigned to a cluster. The fourth constraint ensures that the decision variables are integer.

\subsubsection{\textbf{(3) Vectorizing the Clustering optimization model}}
For the objective function as shown in Equation~\eqref{eq6}, when the $C_p$ is initialized or calculated during iteration, the $||\alpha_i - C_p||$ can be calculated as a constant vector $ \bm{D} \in R^{nm}$. The $y_i$ and $z_i$ comprise column vectors $\bm{Y}$ and $\bm{Z}$, where $ \bm{Y} \in R^{n\cdot m}$ and $ \bm{Z} \in R^{n\cdot m}$. The vectorizing process for $||\alpha_i - C_p||$, $y_i$, and $z_i$ are shown in Equation~\eqref{eq11}-\eqref{eq13}. Therefore, the objective function can be expressed as $\min_{\bm{Z}} \bm{DYZ}$.

\begin{equation}
\centering
\begin{split}
    \bm{d}_1 & = (d_1, d_1,\cdots, d_1)_m \\
    \bm{d}_2 & = (d_2, d_2,\cdots, d_2)_m\\
    & \cdots\\
    \bm{d}_n & = (d_n, d_n,\cdots, d_n)_m \\
    \bm{D} & = (\bm{d}_1, \bm{d}_2, \cdots, \bm{d}_n)_n
\end{split}
\label{eq11}
\end{equation}
where $d_i$ is the distance between the AOI $i$ and its  nearest clustering center.

\begin{equation}
\centering
\begin{split}
    \bm{y}_1 & = (y_1, y_1,\cdots, y_1)_m \\
    \bm{y}_2 & = (y_2, y_2,\cdots, y_2)_m \\
    & \cdots\\
    \bm{y}_n & = (y_n, y_n,\cdots, y_n)_m \\
    \bm{Y} & = (\bm{y}_1, \bm{y}_2, \cdots, \bm{y}_n)_n
\end{split}
\label{eq12}
\end{equation}
where $y_i$ is the predicted order number in the next week for AOI $i$.

\begin{equation}
\centering
\begin{split}
     \bm{Z}=(z_{1,1},z_{1,2},&\cdots,z_{1,m},\\
    z_{2,1},z_{2,2},&\cdots,z_{2,m},\\
    &\cdots,\\
    z_{n,1},z_{n,2},&\cdots,z_{n,m})^T
\end{split}
\label{eq13}
\end{equation}
where $z_{n,m}$ is the decision variable denoting the probability that the AOI $n$ belongs to the cluster $m$.

For the first and second inequality constraints as shown in Equation~\eqref{eq7}-\eqref{eq8}, they are combined as a single inequality constraints $\bm{GZ} \leq \bm{h}$, where $\bm{G}$ and $\bm{h}$ are shown in Equation~\eqref{eq14}-\eqref{eq16}.

\begin{equation}
	\bm{g}_i = \left(
	  \begin{array}{cccc}
	    y_i & 0 & \cdots & 0\\
	    0 &  y_i & \cdots & 0\\
	    \vdots & \vdots & \ddots & \vdots \\
	    0 & 0 & \cdots &  y_i
	  \end{array}
	\right)_{m\times m}
\label{eq14}
\end{equation}

\begin{equation}
	\bm{G} = \left(
	\begin{array}{cccc}
	g_1 & g_2 & \cdots & g_n \\
	-g_1 & -g_2 & \cdots & -g_n
	\end{array}
	\right)_{2m\times mn}
\label{eq15}
\end{equation}

\begin{equation}
\centering
\begin{split}
    \bm{a} & = (a_u, a_u,\cdots, a_u)_m \\
    \bm{b} & = (-b_l, -b_l,\cdots, -b_l)_m\\
    \bm{h} & = \left ( \bm{a}, \bm{b}	\right )^T_{2m}
\end{split}
\label{eq16}
\end{equation}

For the third equality constraints, it can be expressed as the matrix form $\bm{AZ} = \bm{J}$ with the help of identity matrix as shown in Equation~\eqref{eq17}-\eqref{eq18}. $\bm{J} \in R^{mn}$ is an all-ones matrix. The fourth constraint $z_{i,p} \in \{0,1\}$ is slacked into continuous variables $\bm{Z} \in \left [0,1\right ]$.

\begin{equation}
	\bm{a} = \left(
	  \begin{array}{cccc}
	    1 & 0 & \cdots & 0\\
	    0 & 1 & \cdots & 0\\
	    \vdots & \vdots & \ddots & \vdots \\
	    0 & 0 & \cdots & 1
	  \end{array}
	\right)_{n\times n}
\label{eq17}
\end{equation}

\begin{equation}
\begin{split}
  &\bm{A} = (\bm{a}, \bm{a},\cdots, \bm{a})\\
	&= \left(
	  \begin{array}{cccc|c|cccc}
	    1 & 0 & \cdots & 0 &  \cdots & 1 & 0 & \cdots & 0\\
	    0 & 1 & \cdots & 0 & \cdots & 0 & 1 & \cdots & 0\\
	    \vdots & \vdots & \ddots & \vdots & \cdots &\vdots & \vdots & \ddots & \vdots \\
	    0 & 0 & \cdots & 1 & \cdots & 0 & 0 & \cdots & 1
	  \end{array}
	\right)_{n\times mn} 
\end{split}
\label{eq18}
\end{equation}

After the objective function and the constraints are vectorized, the optimization problem can be expressed as Equation~\eqref{eq19}.

\begin{align}
    \min_{\bm{Z}} &\quad (\bm{D} \odot \bm{Y})^T \bm{Z} \label{eq19}\\
    s.t. &\quad \bm{GZ} \leq \bm{h} \label{eq20}\\
         &\quad \bm{AZ} = \bm{J} \label{eq21}\\
         &\quad \bm{Z} \in \left [0,1\right ] \label{eq22}
\end{align}

During the forward pass, the final clustering centers, clustering results, and the modularity loss function can be obtained using Algorithm~\ref{alg:1}. The clustering results are round to a discrete solution using hard maximum to obtain the final clustering results. The modularity loss function is used to evaluate the model performance.

\begin{algorithm}[h]
    \caption{Constrained \textit{K}-means for the optimization layer}
    \label{alg:1}
    \KwIn{The predicted order number $y_i$, the centers of AOIs $\alpha_i$, initialized clustering center $C_{init}$.}
    \KwOut{The clustering results $Z_{new}$, the new clustering center $C_{new}$, the modularity $Q$.}
    \textbf{Initialize}: $C_{old}$, threshold, max iteration number $k_max$ \;
    \For{$k=1; k \le k_max; k++$}{
    Obtain vectorized $\bm{D, Y, Z, G, h, A, J}$ using Equation~\eqref{eq11}-\eqref{eq18} \;
    Solve the vectorized clustering optimization model to obtain $Z_{new}$ and $C_{new}$ using Equation~\eqref{eq19}-\eqref{eq22} \;
    \eIf{$|| C_{old} – C_{new} ||^2 \le threshold$}{
        $C_{old} \leftarrow C_{new}$\;
        $Z_{old} \leftarrow Z_{new}$\;
        break \;
        }{
        continue\;
        }
    }
    Obtain vectorized $\bm{D}, \bm{Y}, \bm{Z}, \bm{G}, \bm{h}, \bm{A}, \bm{J}$ using Equation~\eqref{eq11}-\eqref{eq18} \;
    Solve the vectorized clustering optimization model to obtain the final $Z_{new}$ and $C_{new}$ using Equation~\eqref{eq19}-\eqref{eq22}\;
    Calculate $Q$ using $Z_{new}$ and $C_{new}$ \;
    Output $Z_{new}$, $C_{new}$, and $Q$.
\end{algorithm}

\subsection{Backward pass}

During the backward pass, the goal is to propagate the loss backward through the optimization layer and the prediction layer to optimize the parameters of the prediction layer. The loss function of the PTOClster is the opposite number of the modularity. The original form and matrix form of the modularity are shown in Equation~\eqref{eq23}-\eqref{eq24} \citep{newman2006modularity, wilder2019end}.  
\begin{align}
    Q & =\frac{1}{2m}\sum_{ij}(A_{ij}-\frac{k_ik_j}{2m})\delta(c_i, c_j) \label{eq23}\\
    \bm{Q} & = \frac{1}{2m}Tr(\bm{Z}^T \bm{BZ}) \label{eq24}
\end{align}
where $\bm{Z} \in R_{N \times P}$ is the clustering results. $B_{ij}$ is the modularity matrix with entries as $B_{ij}=A_{ij}-\frac{k_ik_j}{2m}$. $A_{ij}$ is 1 if there is links between node $i$ and $j$ and zero otherwise. $k_i$ and $k_j$ are the degree of node $i$ and $j$. m is the total link numbers. $\delta(c_i, c_j)$ is 1 if node $i$ and $j$ are assigned to the same cluster and zero otherwise.

The technical challenge for the backward pass is to compute the gradient of modularity $\bm{Q}$ with respect to the parameters $\bm{\theta}$ in the prediction layer using the chain rule, as shown in Equation~\eqref{eq25}. Every terms in the right hand side of can be calculated using techniques from matrix differential calculus.

\begin{equation}
    \frac{\partial \bm{Q}}{\partial \bm{\theta}}=\frac{\partial \bm{Q}}{\partial \bm{Z}^{\ast}} \cdot \frac{\partial \bm{Z}^{\ast}}{\partial \bm{Y}} \cdot \frac{\partial \bm{Y}}{\partial \bm{\theta}}
    \label{eq25}
\end{equation}

(1) In terms of $\frac{\partial \bm{Q}}{\partial \bm{Z}^{\ast}}$, it can be obtained through \eqref{eq24} using matrix differential calculus as shown in Equation~\eqref{eq26} 

\begin{equation}
    \frac{\partial \bm{Q}}{\partial \bm{Z}^{\ast}}=\frac{1}{2m} \cdot (\bm{BZ}^{\ast} + \bm{B}^T\bm{Z}^{\ast})
    \label{eq26}
\end{equation}

(2) To obtain the $\frac{\partial \bm{Z}^{\ast}}{\partial \bm{Y}}$, we need to calculate the Jacobian of the optimal solution with respect to the parameters $\bm{Y}$ using the KKT optimality conditions and implicit function theorem \citep{amos2017optnet}. The Lagrangian of Equation~\eqref{eq19}-\eqref{eq22} is shown as Equation~\eqref{eq27}.
\begin{equation}
\begin{split}
    L(\bm{Z}, \bm{\lambda}, \bm{\nu}) = & (\bm{D} \odot \bm{Y})^T \bm{Z} + \\
    & \bm{\nu}^T(\bm{AZ}-\bm{J}) +  \\
    & \bm{\lambda}^T(\bm{GZ}-\bm{h})
\end{split}
\label{eq27}
\end{equation}
where $\bm{\nu}\in R^{2m}$ and $\bm{\lambda} \in R^{n}$ are the dual variables for the equality and inequality constraints, respectively, namely the Lagrangian multipliers. The matrix representation of the KKT conditions for the stationarity (Equation~\eqref{eq28}), primal feasibility (Equation~\eqref{eq29}-\eqref{eq30}), dual feasibility (Equation~\eqref{eq31}), and complementary slackness (Equation~\eqref{eq32}) are shown as follows.

\begin{align}
    \bm{D} \odot \bm{Y} + \bm{A}^T\bm{\nu} + \bm{G}^T\bm{\lambda} & = 0 \label{eq28}\\
    \bm{AZ} - \bm{J} & = 0 \label{eq29}\\
    \bm{GZ} - \bm{h} & \le 0 \label{eq30}\\
    \bm{\lambda} & \ge 0 \label{eq31} \\
    Diag(\bm{\lambda})(\bm{GZ} - \bm{h}) & = 0 \label{eq32}
\end{align}
 where the $Diag(\cdot)$ creates a diagonal matrix from a vector. The equality equations in KKT conditions are used to obtain the differentials as shown in Equation~\eqref{eq33}-\eqref{eq35}, which can be expressed in the form of matrix multiplication by transposition as shown in Equation~\eqref{eq36}.
 
 
\begin{align}
    \begin{split}
         \bm{D} \odot d\bm{Y} + d\bm{D} \odot \bm{Y} + d\bm{A}^T\bm{\nu} + \bm{A}^Td\bm{\nu} + &\\
         d\bm{G}^T\bm{\lambda} + \bm{G}^Td\bm{\lambda}  & = 0 
    \end{split} \label{eq33} \\
     d\bm{AZ} + \bm{A}d\bm{Z}  - d\bm{J}  & = 0  \label{eq34}\\
    \begin{split}
         Diag(\bm{GZ} - \bm{h})d\bm{\lambda} + &\\
         Diag(\bm{\lambda})(d\bm{GZ} + \bm{G}d\bm{Z} - d\bm{h})  & = 0 \label{eq35}
    \end{split}
\end{align}

\begin{align}
\begin{split}
& \left ( \begin{array}{ccc}
0                         & \bm{G}^T             & \bm{A}^T\\
Diag(\bm{\lambda})\bm{G}  & Diag(\bm{GZ}-\bm{h}) & 0\\
\bm{A}                    & 0                    & 0
\end{array} \right) 
\left ( \begin{array}{c}
d\bm{Z} \\
d\bm{ \lambda}  \\
d\bm{\nu} 
\end{array} \right) =  \\ 
& - \left (
\begin{array}{c}
 \bm{D} \odot d\bm{Y} + d\bm{D} \odot \bm{Y} + d\bm{A}^T\bm{\nu} + d\bm{G}^T\bm{\lambda} \\
 Diag(\bm{\lambda} )d\bm{GZ} - diag(\bm{\lambda})d\bm{h}\\
d\bm{AZ} - d\bm{J}
\end{array} \right)
\end{split}
\label{eq36}
\end{align}

When computing the $\frac{\partial \bm{Z}^{\ast}}{\partial \bm{Y}}$, we can simply substitute $d\bm{Y}=\bm{I}$ and set all other differentials in the right hand side to zero in Equation~\eqref{eq36}. Then the $d\bm{Z}$ in the solution of Equation~\eqref{eq36} is just what we want.

(3) As for the $\frac{\partial \bm{Y}}{\partial \bm{\theta}}$, it is the gradient of the prediction results with respect to the parameter $\theta$ in the prediction layer. It can be obtained via the  automatic differentiation mechanics in the PyTorch or TensorFlow.

\section{Experimental results} \label{Experimental}

\subsection{Data description}
In this study, we apply two datasets to evaluate our model. The first dataset contains 35 AOIs with adjacent matrix and the location of AOI centers from Shenzhen, China. The second dataset contains 100 AOIs with adjacent matrix and the location of AOI centers from Hangzhou, China. The corresponding pick-up order numbers of all AOIs are aggregated by one week from January 2019 to April 2021, with a total 116 weeks.

\subsection{Model configuration}
\begin{figure}
    \centering
    \includegraphics[width=0.5\textwidth]{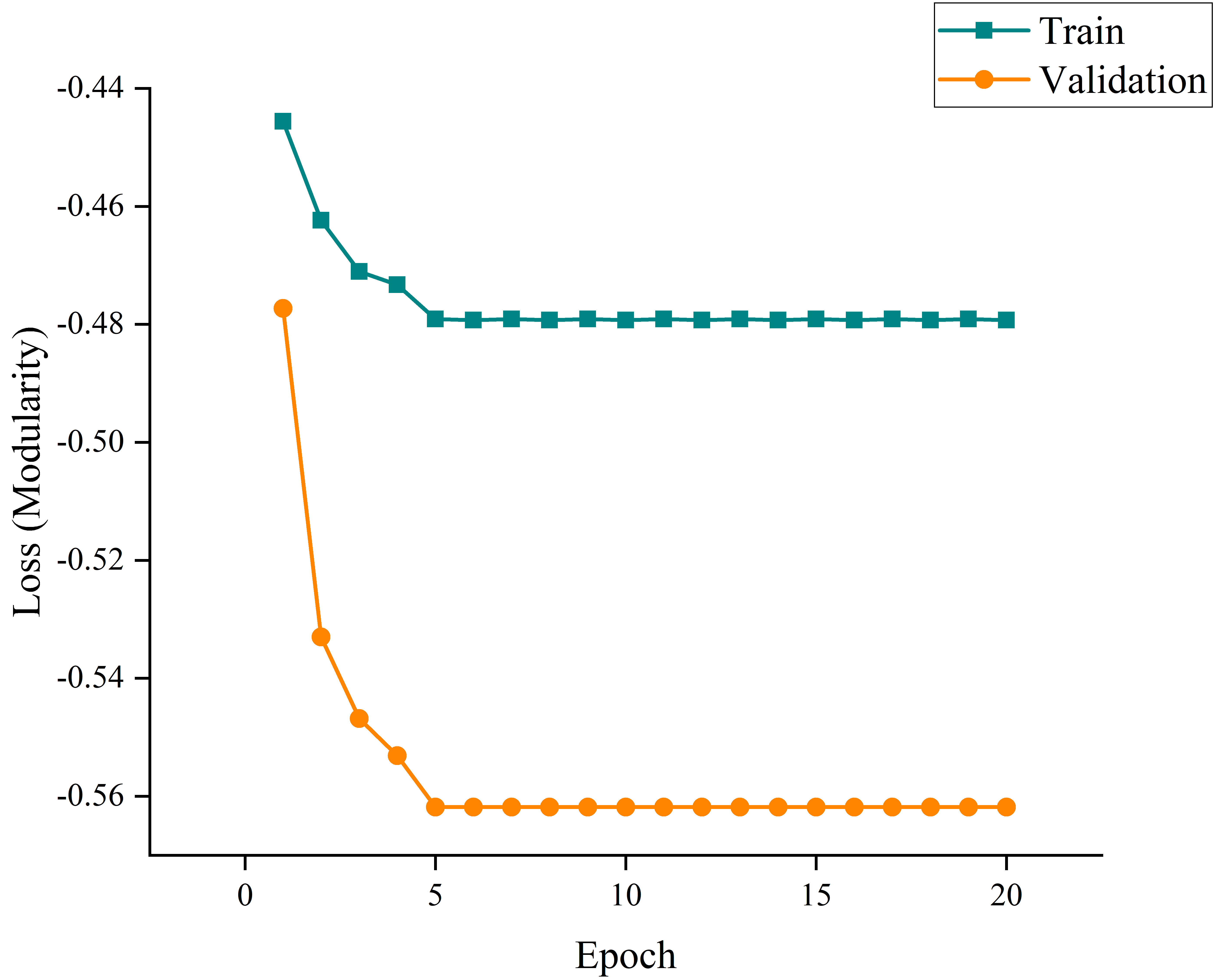}
    \caption{Variation of the training and validating loss for the PTOCluster model}
    \label{fig:loss}
\end{figure}

For the prediction model in the two-stage model and the PTOCluster model, the detailed configuration is shown in Table~\ref{table:1}. Note that we first train an optimal prediction model with the learning rate as 0.001 and then use it to run the two-stage model and train the PTOCluster model with the learning rate as 0.00001. The Loss function for the prediction model is the mean-squared error (MSE). The metrics used to evaluate prediction results are the MSE, root-mean-square error (RMSE), weighted-mean-absolute-percentage error (WMAPE), and $R^2$. The loss function for the end-to-end PTOCluster is modularity. The variation of the training and validating loss of the PTOCluster is shown in Figure~\ref{fig:loss}, showing the stability of the proposed model.

\begin{table}
\caption{Model configuration}
\label{table:1}
\begin{tabular}{|l|cl|}
\hline
\textbf{Parameters}       & \multicolumn{1}{c|}{\textbf{Dataset 1}} & \textbf{Dataset 2}                \\ \hline
AOI number                & \multicolumn{1}{c|}{35 AOIs}            & 100 AOIs                          \\ \hline
time step                 & \multicolumn{1}{c|}{10}                 & 15                                \\ \hline
cluster number            & \multicolumn{1}{c|}{5/6/7/8}            & 5/8/10/15                         \\ \hline
kernel of GCN             & \multicolumn{1}{c|}{10}                 & 15                                \\ \hline
neures of three FC layers & \multicolumn{1}{c|}{1024/512/35}        & \multicolumn{1}{c|}{1024/512/100} \\ \hline
filters of CNN            & \multicolumn{2}{c|}{8}                                                      \\ \hline
kernel of CNN             & \multicolumn{2}{c|}{3×3}                                                    \\ \hline
lr(prediction model)      & \multicolumn{2}{c|}{1e-3}                                                   \\ \hline
lr(PTOCluster)            & \multicolumn{2}{c|}{1e-5 or 1e-6}                                           \\ \hline
time step                 & \multicolumn{2}{c|}{10}                                                     \\ \hline
time span                 & \multicolumn{2}{c|}{Jan, 2019-Apr, 2021, 117 weeks}                         \\ \hline
train sample rate         & \multicolumn{2}{c|}{0.7}                                                    \\ \hline
validation sample rate    & \multicolumn{2}{c|}{0.1}                                                    \\ \hline
test sample rate          & \multicolumn{2}{c|}{0.2}                                                    \\ \hline
lower bound               & \multicolumn{2}{c|}{average order * 0.7}                                    \\ \hline
upper bound               & \multicolumn{2}{c|}{average order * 1.3}                                    \\ \hline
threshold                 & \multicolumn{2}{c|}{2km}                                                    \\ \hline
max iteration number      & \multicolumn{2}{c|}{5}                                                      \\ \hline
initial clustering center & \multicolumn{2}{c|}{sklearn}                                                \\ \hline
\end{tabular}
\end{table}

\subsection{Result analysis}
In this section, we first analyze the prediction results of the two-stage model and the PTOCluster model. The optimal prediction model is used to obtain the optimization results of the two-stage model and to train the PTOCluster model. Therefore, the optimization results are subsequently discussed.

\subsubsection{(1) Prediction results for order number}

\begin{figure}
    \centering
    \includegraphics[width=0.5\textwidth]{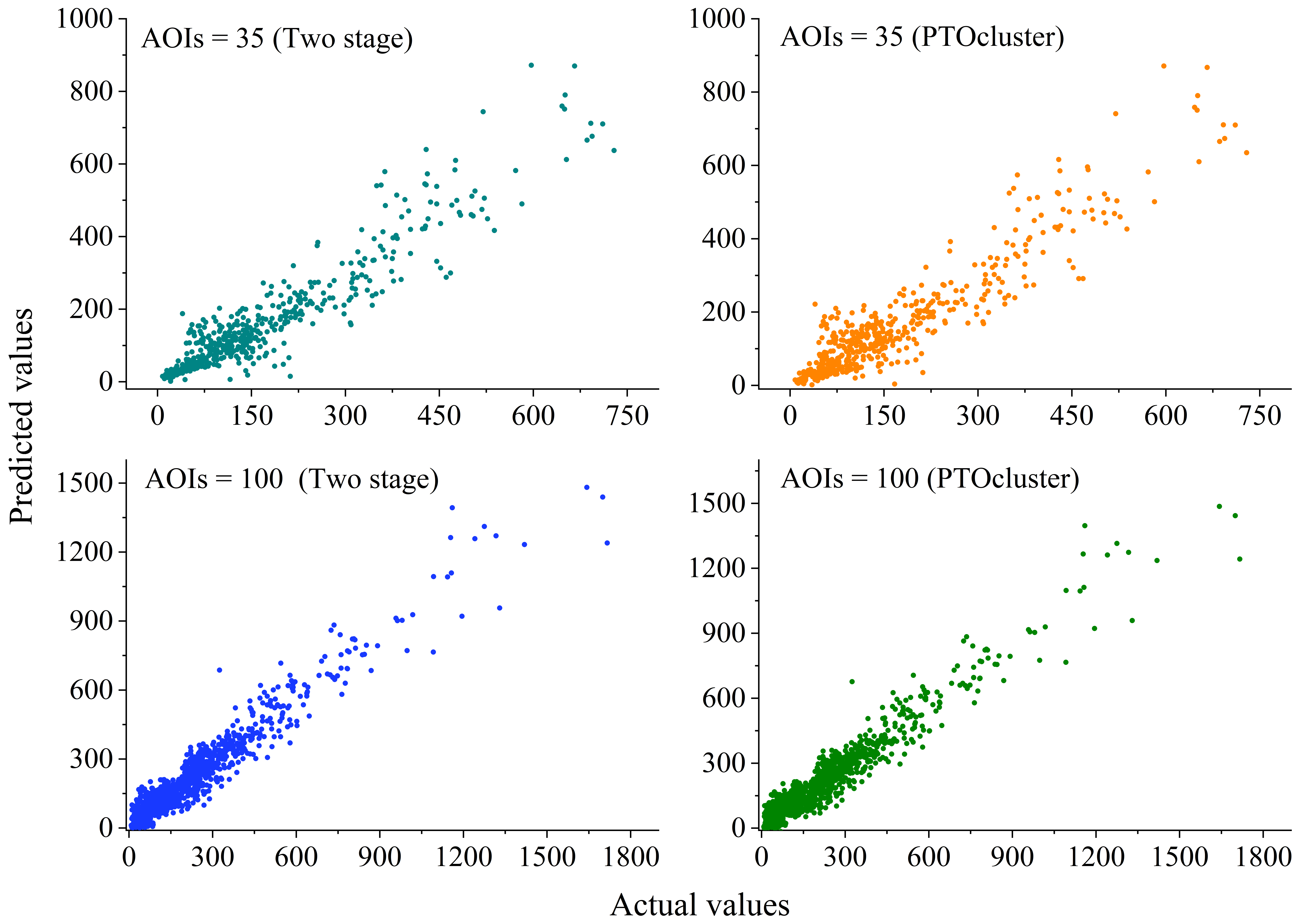}
    \caption{Comparison between actual and predicted values for the two-stage and PTOCluster model}
    \label{fig:predictionresult}
\end{figure}

The metrics evaluating the prediction results for the two-stage model and PTOCluster model are shown in Table~\ref{table:2}. In the two-stage model, we can obtain the optimal prediction results by trial and error. In the fine-tuning process when conducting the PTOCluster model, the prediction accuracy is all slightly affected because the end-to-end model targets improving the optimization or clustering results rather than improving the prediction accuracy.  It indicates that the optimal prediction results might not guarantee the optimal optimization results. It is noted that in the PTOCluster model, the prediction accuracy is not significantly affected. The comparison between actual and predicted values for the two-stage and PTOCluster model is shown in Figure~\ref{fig:predictionresult}. That is because we choose a smaller learning rate (1e-5 or 1e-6) when training the PTOCluster model to guarantee an acceptable prediction accuracy. Therefore, the PTOCluster can trade off between the optimization results and the prediction results, that is the PTOCluster can simultaneously obtain higher optimization results while guarantee an acceptable prediction accuracy.

\begin{table*}[h]
\centering
\caption{Evaluation metrics of the prediction results}
\label{table:2}
\begin{tabular}{|l|cccc|cccc|}
\hline
Dataset                           & \multicolumn{4}{c|}{35   AOIs}                                                                 & \multicolumn{4}{c|}{100   AOIs}                                                                \\ \hline
Criteria                          & \multicolumn{1}{c|}{RMSE}   & \multicolumn{1}{c|}{MAE}    & \multicolumn{1}{c|}{WMAPE} & R2    & \multicolumn{1}{c|}{RMSE}   & \multicolumn{1}{c|}{MAE}    & \multicolumn{1}{c|}{WMAPE} & R2    \\ \hline
Two-stage                         & \multicolumn{1}{c|}{58.004} & \multicolumn{1}{c|}{39.712} & \multicolumn{1}{c|}{0.225} & 0.831 & \multicolumn{1}{c|}{54.456} & \multicolumn{1}{c|}{37.554} & \multicolumn{1}{c|}{0.204} & 0.932 \\ \hline
PTOCluster(Cluster number = 5/5)  & \multicolumn{1}{c|}{58.948} & \multicolumn{1}{c|}{40.989} & \multicolumn{1}{c|}{0.232} & 0.826 & \multicolumn{1}{c|}{56.969} & \multicolumn{1}{c|}{39.743} & \multicolumn{1}{c|}{0.216} & 0.925 \\ \hline
PTOCluster(Cluster number = 6/8)  & \multicolumn{1}{c|}{61.153} & \multicolumn{1}{c|}{41.085} & \multicolumn{1}{c|}{0.233} & 0.812 & \multicolumn{1}{c|}{55.586} & \multicolumn{1}{c|}{38.543} & \multicolumn{1}{c|}{0.210} & 0.929 \\ \hline
PTOCluster(Cluster number = 7/10) & \multicolumn{1}{c|}{60.039} & \multicolumn{1}{c|}{41.760} & \multicolumn{1}{c|}{0.237} & 0.819 & \multicolumn{1}{c|}{55.748} & \multicolumn{1}{c|}{40.540} & \multicolumn{1}{c|}{0.221} & 0.927 \\ \hline
PTOCluster(Cluster number = 8/15) & \multicolumn{1}{c|}{58.918} & \multicolumn{1}{c|}{41.855} & \multicolumn{1}{c|}{0.237} & 0.826 & \multicolumn{1}{c|}{54.575} & \multicolumn{1}{c|}{37.968} & \multicolumn{1}{c|}{0.207} & 0.932 \\ \hline
\end{tabular}
\end{table*}

\begin{table*}[t]
\centering
\caption{Modularity in the optimization results}
\label{table:3}
\begin{tabular}{|l|cccc|cccc|}
\hline
Dataset        & \multicolumn{4}{c|}{36 AOIs}                                                                      & \multicolumn{4}{c|}{100 AOIs}                                                                    \\ \hline
Cluster number & \multicolumn{1}{c|}{5}      & \multicolumn{1}{c|}{6}       & \multicolumn{1}{c|}{7}      & 8      & \multicolumn{1}{c|}{5}      & \multicolumn{1}{c|}{8}      & \multicolumn{1}{c|}{10}     & 15     \\ \hline
Two-stage      & \multicolumn{1}{c|}{0.539}  & \multicolumn{1}{c|}{0.510}   & \multicolumn{1}{c|}{0.479}  & 0.464  & \multicolumn{1}{c|}{0.596}  & \multicolumn{1}{c|}{0.630}  & \multicolumn{1}{c|}{0.605}  & 0.547  \\ \hline
PTOCluster     & \multicolumn{1}{c|}{0.580}  & \multicolumn{1}{c|}{0.562}   & \multicolumn{1}{c|}{0.522}  & 0.493  & \multicolumn{1}{c|}{0.623}  & \multicolumn{1}{c|}{0.647}  & \multicolumn{1}{c|}{0.627}  & 0.553  \\ \hline
Improvement    & \multicolumn{1}{c|}{7.57\%} & \multicolumn{1}{c|}{10.19\%} & \multicolumn{1}{c|}{9.02\%} & 6.35\% & \multicolumn{1}{c|}{4.61\%} & \multicolumn{1}{c|}{2.68\%} & \multicolumn{1}{c|}{3.69\%} & 1.20\% \\ \hline
\end{tabular}
\end{table*}

\subsubsection{(2)Optimization results for the modularity}
The optimization results are shown in Table~\ref{table:3}. The higher the modularity, the better the clustering results are. As is shown, the modularity of the PTOCluster is all improved compared with the two-stage model. That means, considering the order number constraints and the centers of AOIs, better clustering results are obtained so that the AOIs can be better assigned to couriers according to the clustering results. The improvements for the two datasets range from 1.20\% to 10.09\%, showing the effectiveness of the end-to-end framework. 

\begin{figure}
    \centering
    \includegraphics[width=0.48\textwidth]{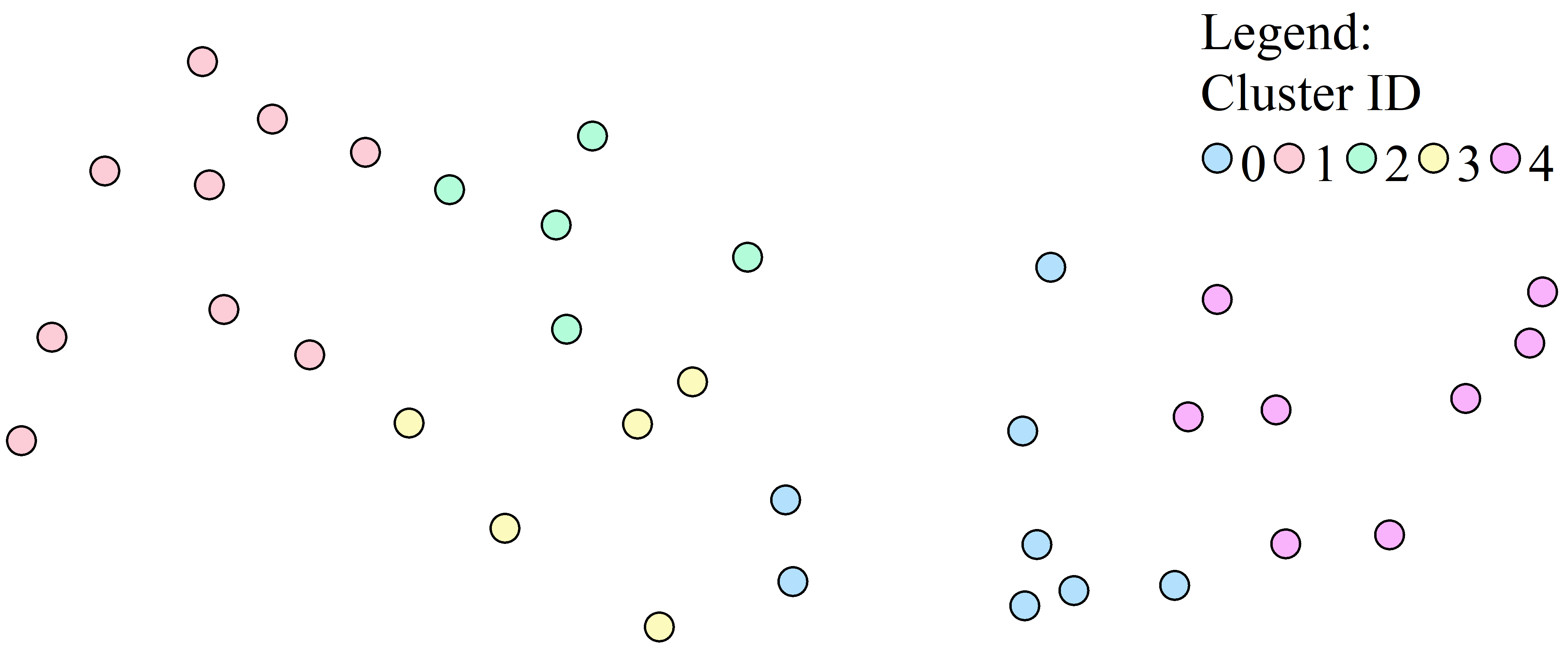}
    \caption{An example of the clustering result for the 35 AOIs}
    \label{fig:cluster35}
\end{figure}

\begin{figure}
    \centering
    \includegraphics[width=0.48\textwidth]{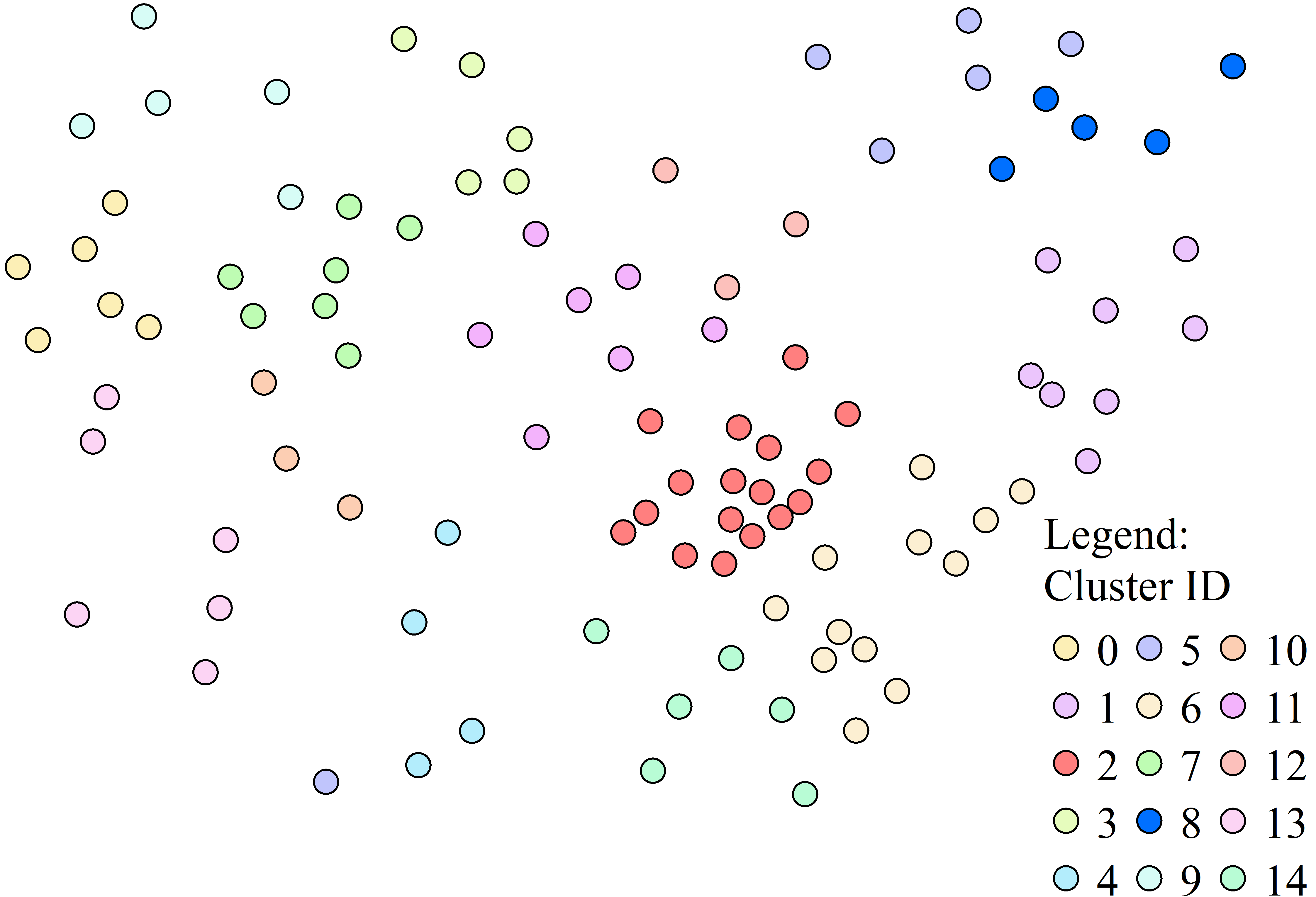}
    \caption{An example of the clustering result for the 100 AOIs}
    \label{fig:cluster100}
\end{figure}

The 5 clustering results for the 35 AOIs and 15 clustering results for the 100 AOIs are shown in Figure~\ref{fig:cluster35} and \ref{fig:cluster100}, respectively. As is shown, the clustering results are reasonable geographically, and adjacent AOIs are assigned into one cluster, indicating that the proposed differential constrained \textit{K}-means optimization model is effective.

\section{Conclusion} \label{Conclusion}
This paper proposes an end-to-end PTOCluster framework to dynamically cluster AOIs and assign AOIs to couriers in Express systems. First, in the prediction part, a GCN-based prediction model is proposed to predict the AOIs in the next week. Second, In the optimization part, we propose a differential constrained \textit{K}-means clustering model as an optimization layer to conduct the assignment model. Theoretically, this is the first time that a differential constrained \textit{K}-means optimization layer applied in deep learning architecture is proposed. Finally, the prediction part and the optimization part are embedded into an end-to-end framework, which can directly obtain a better final solution from original data. The main conclusions are summarized as follows.
\begin{itemize}
    \item The end-to-end PTOCluster is proved to be effective for improving the clustering results. The improvements compared with conventional two-stage methods range from 1.20\% to 10.09\%.
    \item The proposed differential constrained \textit{K}-means optimization layer is feasible to be embedded into deep learning architecture.
    \item The accuracy of the proposed GCN-based deep learning model for order prediction is favorable.
    \item The proposed PTOCluster can be used to intelligently and dynamically conduct AOI assignments for couriers in express systems.
\end{itemize}

\section{Acknowledgement}
We wish to thank the anonymous reviewers for the valuable comments, suggestions, and discussions. This work was supported by the Fundamental Research Funds for the Central Universities (No. 2021RC270) and the National Natural Science Foundation of China (Nos. 71621001, 71825004).

\bibliographystyle{named}
\bibliography{ijcai22}
\end{document}